\documentclass{article}
\usepackage{ijcai20}

\usepackage{times}

\usepackage{soul}
\usepackage{url}
\usepackage[hidelinks]{hyperref}
\usepackage[utf8]{inputenc}
\usepackage[small]{caption}
\usepackage{graphicx}
\usepackage{amsmath}
\usepackage{booktabs}

\usepackage{color}

\usepackage{amsthm}

\usepackage{algorithm}
\usepackage{algorithmic}
\usepackage{multirow}
\usepackage{multicol}

\usepackage{epsfig}
\usepackage{amsmath}
\usepackage{amssymb}
\usepackage{subfigure} 
\newcommand{\zhuo}[1]{\textcolor{black}{#1}}
\newcommand{\zhuonew}[1]{\textcolor{black}{#1}}

\title{CP-NAS: Child-Parent Neural Architecture Search for Binary Neural Networks }

\author{
Li'an~Zhuo$^1$\and
Baochang~Zhang$^1$\footnote{Baochang Zhang is the corresponding author.}\and
Hanlin~Chen$^1$\and
Linlin~Yang$^2$\and
Chen Chen$^3$\and \\
Yanjun Zhu$^4$\And
David Doermann$^4$
\\
\affiliations
$^1$School of Automation Science and Electrical Engineering, Beihang University\\
$^2$University of Bonn\\
$^3$University of North Carolina at Charlotte\\
$^4$University at Buffalo\\
\emails
\{lianzhuo, bczhang, hlchen\}@buaa.edu.cn
}

\begin{document}
	
	\maketitle
	
	\begin{abstract}
		Neural architecture search (NAS) proves to  be among the best approaches  for many tasks by generating an application-adaptive neural architecture, which is still challenged by high computational cost and memory consumption.
		At the same time,  1-bit convolutional neural networks (CNNs) with binarized weights and activations show their potential for resource-limited embedded devices. 
		One natural approach is to use 1-bit CNNs to reduce the computation and memory cost of NAS by taking advantage of the strengths of each in a unified framework.   To this end, a Child-Parent (CP) model is introduced to a differentiable NAS to search the binarized architecture (Child) under the supervision of a full-precision model (Parent).  
		In the search stage, the Child-Parent model uses an indicator generated by the child and parent model accuracy to evaluate the performance and abandon operations with less potential. In the training stage, a kernel-level CP loss is introduced to optimize the binarized network. 
		Extensive experiments demonstrate that the proposed CP-NAS achieves a comparable accuracy with traditional NAS on both the CIFAR and ImageNet databases. It achieves the accuracy of $95.27$\% on CIFAR-10, $64.3$\% on ImageNet with binarized weights and activations, and a $30\%$ faster search  than  prior arts.
	\end{abstract}
	
	\begin{figure*}[htbp!]
		\centering
		\includegraphics[scale=.48]{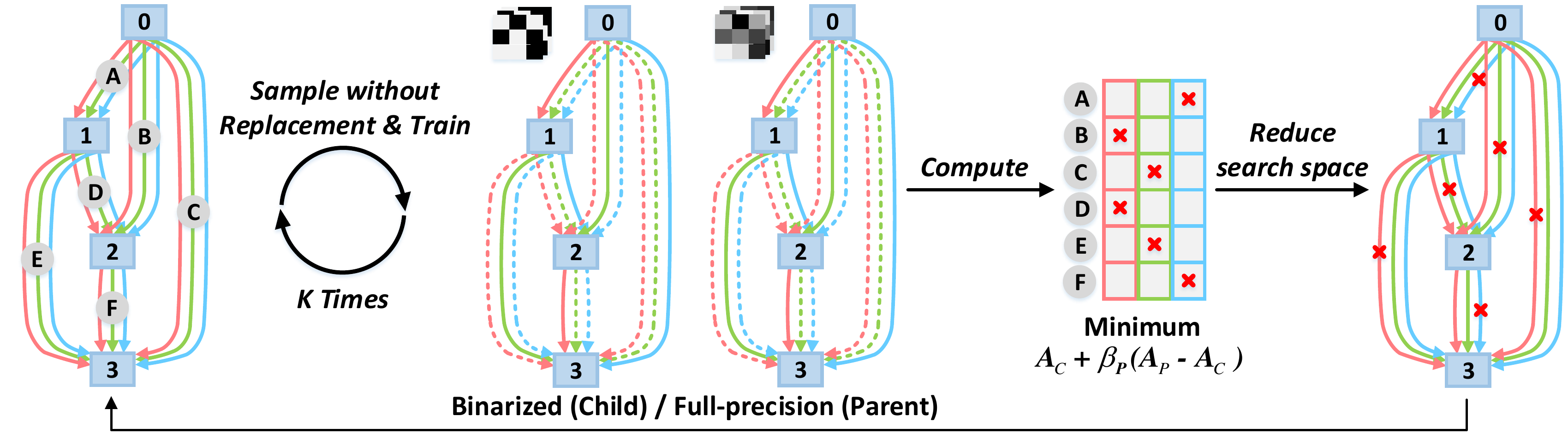}
		\caption{The main framework of the proposed Child-Parent search strategy. In a loop, we first sample the operation without replacement for each edge from the search space, and then train the Child model and Parent model generated by the same architecture simultaneously. Second, we use the Eqs.~\ref{eq:cp} and \ref{eq:soft_performance} to compute the evaluation indicator calculated by the accuracy of both models on the validation dataset. Until all the operations are selected, we remove the operation in each edge with the worst performance.}
		\label{fig:parent-guided}
	\end{figure*}

	\section{Introduction}
	Neural architecture search (NAS) has attracted a great deal of attention with a remarkable performance in many computer vision tasks. The goal is to design network architectures automatically to replace conventional hand-crafted counterparts, \zhuo{but at the expense of} huge search space and  high computational cost. To achieve efficient NAS, one line of existing NAS approaches focus  on improving their search efficiency to  explore the large search spaces, reducing the search time from  thousands of GPU days \cite{Zoph2018CVPR,zoph2016neural}  to few GPU days \cite{cai2018efficient,liu2018darts,xu2019pcdarts,Chen_2019_ICCV}.
	These approaches were also developed into a more elegant framework named one-shot architecture search. Another line  of NAS aims to 
	search a more efficient network. ProxylessNAS \cite{cai2018proxylessnas} introduces latency loss to search architectures on the target task instead of adopting the conventional proxy-based framework. EfficientNet \cite{tan2019efficientnet} introduces a new scaling method that uniformly scales all dimensions of depth/width/resolution using a simple yet highly effective compound coefficient to obtain efficient networks. Binarized neural architecture search (BNAS) \cite{chen2019BNAS} searches binarized networks with a significant memory saving, which provides a more promising way to efficiently find network architectures. However, BNAS only focuses on the kernel binarization, while the extremely compressed 1-bit CNNs with binarized weights and activations have not been well explored in the field of NAS.

	Comparatively speaking, 1-bit CNNs based on hand-craftd architectures have been extensively researched. 
	\zhuonew{Filters binarization has been} used in conventional  CNNs  to compress deep models \cite{rastegari2016xnor,paper10,paper14,paper15}, showing up to 58$\times$ speedup and 32$\times$ memory saving, which is widely considered as one of the most efficient ways to perform computing on embedded devices with low computational cost. 
	In \cite{paper15}, the XNOR network is presented where both the weights and inputs attached to the convolution are approximated with binarized values. This results in an efficient implementation of convolutional operations by reconstructing the unbinarized filters with a single scaling factor. 
	In \cite{gu2018projection}, a projection  convolutional neural network (PCNN) is proposed to implement binarized neural networks (BNNs) based on a simple back propagation algorithm. \cite{zhao2019bonn} proposes Bayesian optimized 1-bit CNNs, taking \zhuo{advantage} of Bayesian learning to significantly improve the performance of extreme 1-bit CNNs. 
	Binarized models show the advantages on computational cost reduction and memory saving, however, they suffer from poor performance  
	in practical applications. 
	{There still remains a gap between 1-bit weights/activations and full-precision counterparts, which motivates us to explore the potential relationship between 1-bit and full-precision models to evalutate \zhuo{the performance of binarized networks} based on NAS.}
	
	In this paper, we introduce a Child-Parent model  to efficiently search a binarized network architecture in a unified framework.
	The search strategy for Child-Parent model consists of three steps shown in Fig.~\ref{fig:parent-guided}. First, we sample the operations without replacement and construct two classes of sub-networks that share the same architecture, i.e., binarized networks (Child) and full-precision networks (Parent). Second, we train both sub-networks and obtain the performance indicator of the corresponding operations by calculating the child network accuracy and the accuracy loss between child and parent networks. It is observed that the \zhuonew{worse} operations in the early stage usually have the worse performance at the end. Based on this observation, we then remove the operation with the worst performance according to the performance indicator. This precoess is repeated until there is only one operation left in each edge. For binarized optimization of Child-Parent model, we reformulate the traditional binarization loss as a kernel-level Child-Parent loss.
	The main contributions of our paper include:
	
	\begin{itemize}
		\item
		A Child-Parent model is introduced to guide the binarized architecture search and to optimize BNNs in a unified framework. 
		\item
		An indicator is proposed to evaluate the operation performance based on Child-Parent model. The search space is greatly reduced through this search strategy for Child-Parent model, 
		which improves the search efficiency significantly.

		\item
		Extensive experiments demonstrate the superiority of the proposed algorithm over other light models on the CIFAR-10 and ImageNet datasets. 
	\end{itemize}

	\section{Child-Parent NAS}
	
	In this section, we first describe the proposed CP-NAS, our Child-Parent model for NAS. Then, 
	the search space and strategy for CP-NAS is introduced to effectively find an powerful binarized architecture. Finally, 
	a kernel-level CP loss is proposed for binarized optimization.  
	The framework of CP-NAS is shown in Fig.~\ref{fig:parent-guided},  and details are provided below.
	
	\begin{figure}[htbp!]
		\centering
		\includegraphics[scale=.32]{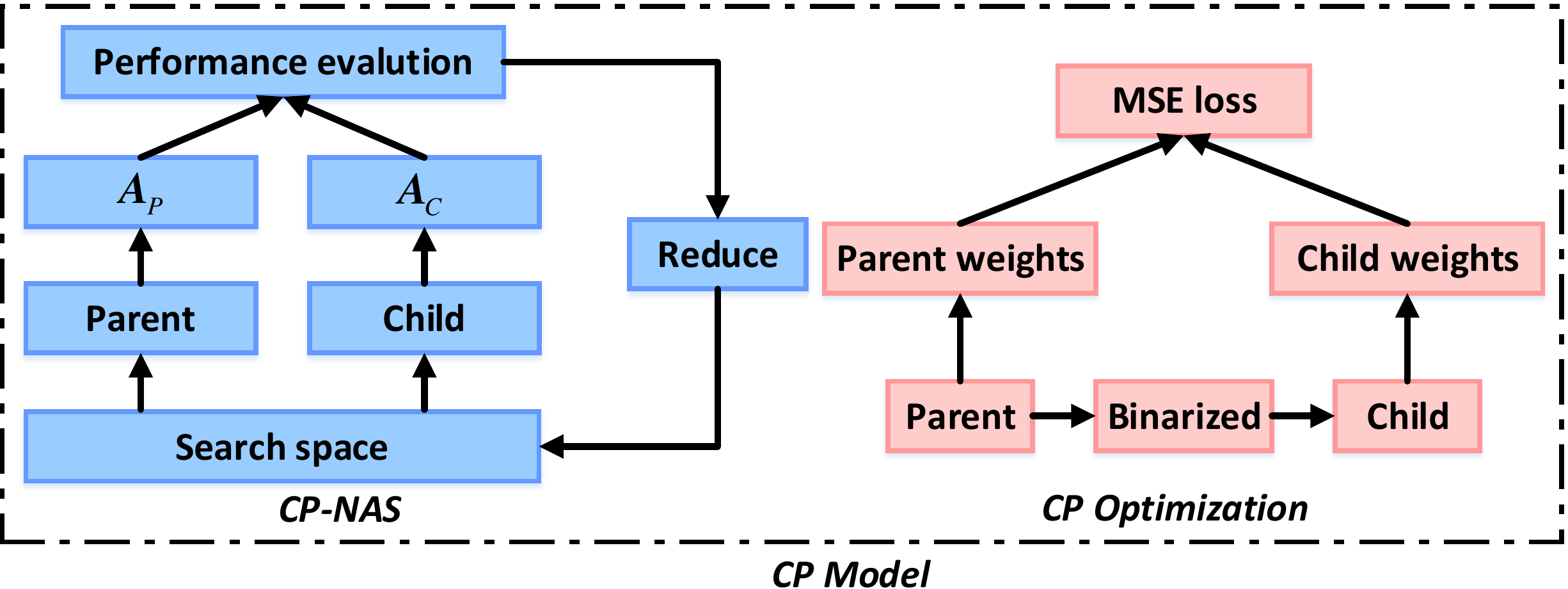}
		\caption{The main framework of Child-Parent model. 
			The Child-Parent model focuses on both the binarized architecture search (left) and binarized optimization (right).}
		\label{fig:cp-model}
	\end{figure}

	\subsection{Child-Parent Model for Network Binarization} 
	\label{sec:PBS}
	
	
	Network binarization, which calculates neural networks with 1-bit weights and activations to fit the  full-precision network, can  significantly compress the deep convolutional nerual networks (CNNs). Prior work \cite{zhao2019bonn} usually investigates the binarization problem by exploring  the full-precision model to guide the optimization of binarized models.  Based on the investigation,   
	we reformulate NAS-based network binarization as a Child-Parent model as shown in Fig \ref{fig:cp-model}. The binarized model and the  full-precision counterpart are the child and parent models respectively.

	Conventional NAS is inefficient  due to the complicated reward computation in network training where the evaluation of a structure is usually done after the network training converges. There are also some methods to  perform the evaluation of a cell during the training of the network. 
	\cite{zheng2019multinomial} points out that the best choice in early stages is  not  necessarily  the final optimal one, however, the worst operation in the early stages usually has a bad performance at the end. And this phenomenon will become more and more \zhuonew{significant} as the training goes. Based on this observation, we propose a simple yet effective operation removing process, which is the key task of the proposed CP model. 
	
\begin{figure}[tbp!]
		\centering
		\includegraphics[scale=.42]{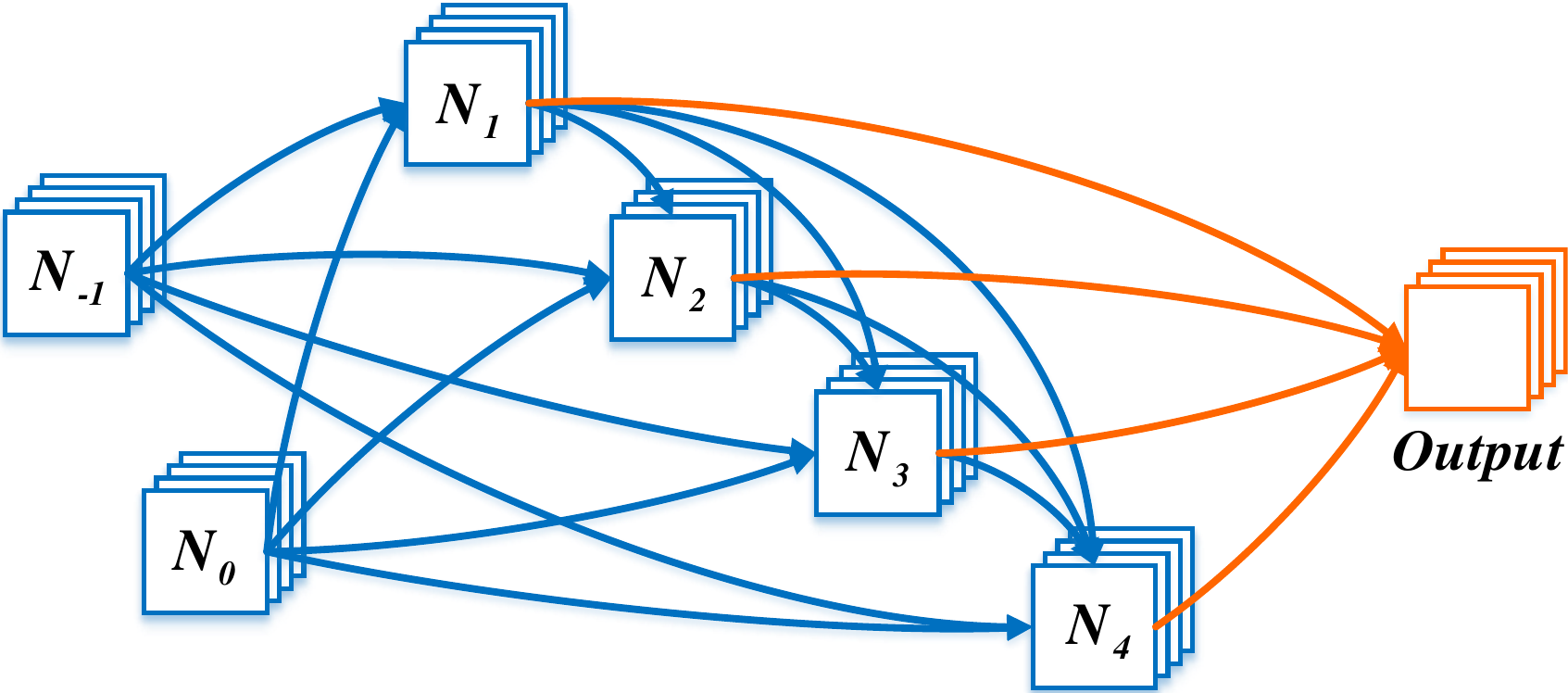}
		\caption{The cell architecture for CP-NAS. One cell includes $2$ input nodes, $4$ intermediate nodes and $14$ edges (blue).}
		\label{fig:cell}
	\end{figure}

	\zhuo{Intuitively, the difference between the children and parents ability, and how much children can  independently handle their problems,  are two main aspects that should be considered to define a reasonable performance evaluation measure.
		Our Child-Parent model introduces a  similar performance indicator to improve the search efficiency.}
	The performance indicator includes two parts,  the performance loss between the binarized network (Child) and the full-precision network (Parent), and  the performance of the binarized network (Child). We can thuse define it for each operation of the sampled network as
	\begin{equation}
	z_{k,t}^{{(i,j)}}= A_{C,t}+\beta_P(A_{P,t}-A_{C,t})
	\label{eq:cp}
	\end{equation}
	where $A_{P,t}$ and $A_{C,t}$ represent the network performance calculated by the accuracy of the full-precision  model (Parent) and the binarized  model (Child) on the validation dataset, and $\beta_P$ is the hyper-parameter to control the performance loss. $i$,$j$ represent the index of the node to generate edge $(i,j)$ shown in Fig.~\ref{fig:cell}, $k$ is the operation index of corresponding edge, and $t$ represents the $t$th sampling process. 
	Note that we use the performance of the sampled network to evaluate the performance of the corresponding selected operations.  
	
	CP-NAS not only uses the accuracy on the validation dataset to guide the search process directly, but also takes the information of full-precision model into consideration to better investigate the full potential that the binarized model can  ultimately reach. Additional details are provided in the following section.
	
	As shown in Fig.~\ref{fig:cp-model}, unlike the traditional teacher-student model \cite{hinton2015distilling}  which transfers the generalization ability of the first model to a smaller model by using the class probabilities as “soft targets”, the Child-Parent model focuses on the  performance measure particularly suitable for NAS-based network binarization. Furthermore,  the loss function for the teacher-student model is constrained on the feature map or the output, while ours focuses on the kernel weights to minimize the variations between two networks.
	
	\subsection{Search Space}
\begin{figure}[tbp!]
		\centering
		\includegraphics[scale=.40]{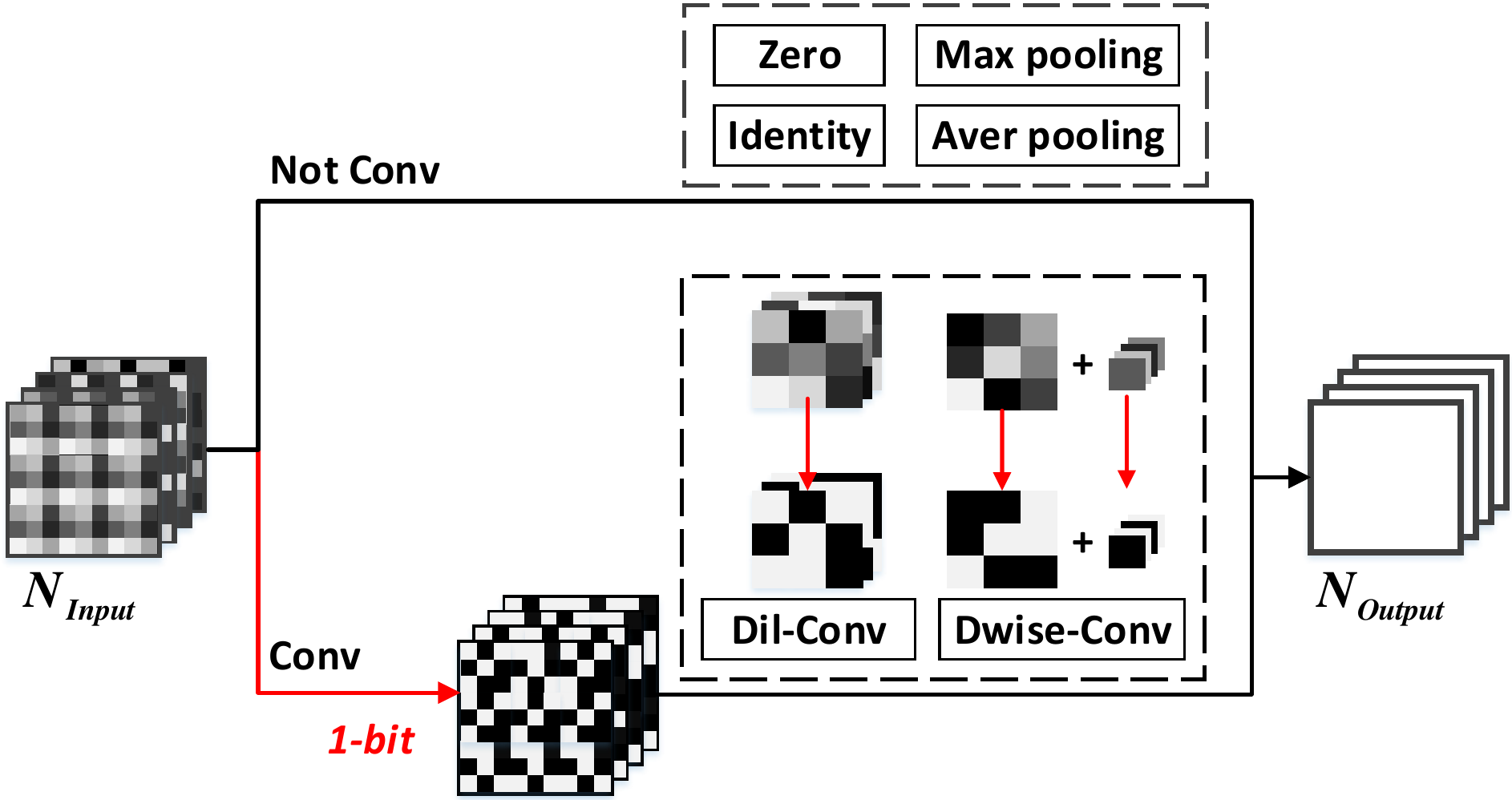}
		\caption{The operations of each edge. Each edge has $4$ convolutional operations, including $2$ types of binarized convolutions with $3*3$ or $5*5$ receptive fields, and $4$ non-convolutional operations.}
		\label{fig:operation}
	\end{figure}

	We search for computation cells as the building blocks of the final architecture. As in \cite{zoph2016neural,Zoph2018CVPR,liu2018darts,real2019regularized}, we construct the network with a pre-defined number of cells and each cell is a fully-connected directed acyclic graph (DAG) $\mathcal{G}$ with $M$ nodes,  $\{N_1, N_2, ..., N_M\}$.  For simplicity, we assume that each cell only takes the outputs of the two previous cells as input and each input node has pre-defined convolutional operations for preprocessing. Each node $N_j$ is obtained by $N_j = \sum_{i<j} o^{(i,j)}(N_i)$. $N_i$ is the dependent node of  $N_j$ with the constraint $i<j$ to avoid cycles in a cell. We also define nodes $N_{-1}$ and $N_{0}$ without inputs as the first two nodes of a cell. 
	Each node is a specific tensor like a feature map, and each directed edge $(i,j)$ denotes an operation $o^{(i,j)} (.)$ shown in Fig.~\ref{fig:operation}, which is sampled from following $K = 8$ operations: 
	\begin{multicols}{2}
		\small
		\begin{itemize}
			\item no connection (zero)
			\item skip connection (identity)
			\item $3\times3$ dilated convolution with rate $2$
			\item $5\times5$ dilated convolution with rate $2$
			\item $3\times3$ max pooling
			\item $3\times3$ average pooling
			\item $3\times3$ depth-wise separable convolution
			\item $5\times5$ depth-wise separable convolution
		\end{itemize}
	\end{multicols}
	
	\zhuo{ We replace the convolution with a binarized form.  
	\zhuo{ We also remove the ReLU  operation to avoid the vanishing of the negative in the 1-bit convolution.} The optimization of BNNs is more challenging than that of the conventional CNNs~\cite{rastegari2016xnor}, \cite{gu2018projection}, since binarization brings additional computation burdens to NAS.}

	\subsection{Search Strategy for CP-NAS}
	
	As shown in Fig.~\ref{fig:parent-guided}, we randomly sample one operation from the $K$ operations in $\mathcal{O}^{(i,j)}$ for every edge and then obtain the performance based on Eq.~\ref{eq:cp} by training the sampled parent and child networks for one epoch. Finally, we assign this performance to all the sampled operations. These steps are performed $K$ times by sampling without replacement, leading to each operation having exactly one accuracy for every edge for fairness.
	
	We repeat the complete sampling process $T$ times. Thus each operation for every edge has $T$ performance \zhuonew{measures} $\{z_{k,1}^{(i,j)}, z_{k,2}^{(i,j)}, ..., z_{k,T}^{(i,j)}\}$ calculated by Eq.~\ref{eq:cp}. 
	Furthermore, to reduce the undesired fluctuation in
	the performance evaluation, we normalize the performance 
	of $K$ operations for each edge to obtain the final evaluation indicator as
	
	\begin{equation}\label{eq:soft_performance}
	e(o^{(i,j)}_k) = \frac{exp\{\bar{z}_k^{(i,j)}\}}{\sum_{k'} exp\{\bar{z}_{k'}^{(i,j)}\}},
	\end{equation}
	where $\bar{z}_k^{(i,j)} = \frac{1}{T}  \sum_t z_{k,t}^{(i,j)}$. Along with the increasing epochs, following \cite{zheng2019multinomial} and \cite{chen2019BNAS}, we progressively abandon the worst evaluation operation from each edge until there is only one operation for each edge. The complete algorithm is shown in Alg.~\ref{alg:algorithm}.
	
	
	\begin{algorithm}[tb]
		\caption{Child-Parent NAS}
		\label{alg:algorithm}
		\textbf{Input}: Training data, Validation data\\
		\textbf{Parameter}: Searching hyper-graph: $\mathcal{G}$, $K=8$, $e(o^{(i,j)}_k)=0$ for all edges\\
		\textbf{Output}: Optimal structure $\alpha$
		\begin{algorithmic}[1] 
			\WHILE{$(K>1)$}
			\FOR{$t= 1,...,T$ \rm epoch}
			\FOR{ $ e=1,...,  K  $ \rm epoch}
			\STATE Select an architecture by sampling (without replacement) one operation from $\mathcal{O}^{(i,j)}$ for every edge; \\
			\STATE Construct the Child model and Parent model with the same selected architecture, and then train both models to get the accuracy on the validation data; \\
			Use Eq.\ref{eq:cp} to compute the performance and assign that to all the sampled operations; \\
			\ENDFOR
			\ENDFOR
			\STATE Update $e(o^{(i,j)}_k)$ using Eq.~\ref{eq:soft_performance}; \\
			\STATE Reduce the search space \{$\mathcal{O}^{(i,j)}$\} with the worst performance evaluation by  $e(o^{(i,j)}_k)$ ;\\
			\STATE $K = K -1$; \\
			\ENDWHILE
			\STATE \textbf{return} solution
		\end{algorithmic}
	\end{algorithm}

	\subsection{Optimization for 1-bit CNNs} 
	
	Inspired by XNOR and PCNN, we reformulate the binarized optimization as Child-Parent optimization in our unified framework. 
	
	To binarize the weights and activations of CNNs, we introduce the kernel-level Child-Parent loss for binarized optimization from two respects. First,  we minimize the distributions between the full-precision filters and their corresponding binarized filters. Second, we minimize the intra-class compactness based on the output features. We then have a loss function  as
	
	\begin{equation}\label{eq_LossM}
	\begin{aligned}
	\mathcal{L}_{\hat{H}}=\sum_{c,l} \text{MSE}
	(H^l_c,\hat{H}^l_c) + \frac{\lambda}{2}\sum_{s}{\Vert f_{C,s}(\hat{H})-\overline{f}_{C,s}(H) \Vert^2},
	\end{aligned}
	\end{equation}
	where  $\lambda$ is a hyperparameter to balance the two terms. $H^l_c$ is the $c$th full-precision filter of the $l$th convolutional layer and $\hat{H}^l_c$ denotes its corresponding reconstructed filter; 
	$\text{MSE}(\cdot)$ represents the mean square error (MSE) loss.  The second term is used to minimize the intra-class compactness, since the binarization process  causes feature variations. $f_{C,s}(\hat{H})$ denotes the feature map of the last convolutional layer for the $s$th sample, and $\overline{f}_{C,s}(\hat{H})$ denotes the class-specific mean feature map for corresponding samples.  Combining $\mathcal{L}_{\hat{H}}$ with the conventional loss $\mathcal{L}_{CE}$, we obtain the final loss as
	\begin{equation}\label{eq_Loss}
	\mathcal{L}=\mathcal{L}_{CE}+\mathcal{L}_{\hat{H}}.
	\end{equation} 
	
	$\mathcal{L}$ and its derivatives are easily calculated by the Pytorch  package directly.

	\section{Experiments}

	\begin{table*}
		\centering
		\normalsize
		\begin{tabular}{lcccccc}
			\toprule[1pt]
			\multirow{2}{*}{\textbf{Architecture}} & \textbf{Test Error} & \textbf{\# Params} & \textbf{W}&\textbf{A}& \textbf{Search Cost} & \textbf{Search} \\
			& \textbf{(\%)} & \textbf{(M)} & & & \textbf{(GPU days)} & \textbf{Method} \\
			\hline
			
			WRN-22 \cite{zagoruyko2016wide}  & 5.04 & 4.33 &32&32 & - & Manual  \\
			DARTS \cite{liu2018darts} & 2.83 & 3.4 &32&32 & 4 & Gradient-based \\%
			PC-DARTS \cite{xu2019pcdarts} & 2.78 & 3.5 &32&32 & 0.15 & Gradient-based \\ 
			\hline
			WRN-22 (PCNN) \cite{gu2018projection} & 5.69 & 4.33&1&32  & - & Manual \\							
			BNAS (PCNN) \cite{chen2019BNAS} & 3.94 & 2.6 &1&32 & 0.09 & Performance-based \\
			BNAS (PCNN, larger) \cite{chen2019BNAS} & 3.47 & 4.6 &1&32 & 0.09  & Performance-based \\ 			
			\hline
			
			WRN-22 (BONN) \cite{zhao2019bonn} & 8.07 & 4.33&1&1  & - & Manual \\				
			BNAS$^\dagger$ & 8.29 & 4.5 &1&1 & 0.09 & Performance-based \\
			\textbf{CP-NAS} (Small) & \textbf{6.5} & 2.9 &1&1 & 0.1 & Child-Parent model \\
			\textbf{CP-NAS} (Medium) & \textbf{5.72} & 4.4 &1&1 & 0.1  & Child-Parent model \\
			\textbf{CP-NAS} (large) & \textbf{4.73} & 10.6 &1&1 & 0.1  & Child-Parent model \\ 					
			\bottomrule[1pt]
		\end{tabular}
		\caption{Test error on CIFAR-10. ’W’ and ’A’ refer to the weight and activation bitwidth respectively. ’\textbf{M}’ means million ($10^6$). \zhuonew{BNAS$^\dagger$ is approximately implemented by us  by setting $\beta_P = 0$ in CP-NAS, which means that we only use the performance measure for the  operation selection.}}
		\label{tab:cifar}
	\end{table*}

	In this section, we compare our CP-NAS with the state-of-the-art NAS methods and 1-bit CNNs methods on two publicly available datasets: CIFAR-10~\cite{krizhevsky2014cifar} and ILSVRC12 ImageNet ~\cite{russakovsky2015imagenet}.

	
	\subsection{Training and Search Details}
	


	In our experiments, we first search binarized  neural architectures on an over-parameterized network on CIFAR-10, and then evaluate the best architecture with a stacked deeper network on the same dataset.  We  perform experiments to search binarized architectures directly on ImageNet. We run the experiment multiple times and find that the resulting architectures  show only a slight variation in performance, which demonstrates the stability of our method.

	We use the same datasets and evaluation metrics as previous NAS works \cite{liu2018darts,cai2018path,Zoph2018CVPR,liu2018progressive}. 
	The color intensities of all images are normalized to $[-1, +1]$. 
	\zhuo{During the architecture search, the training set of the dataset is divided into two subsets, one for training the network weights and the other for perfomrance evaluation as a validation set.}  
	
	
	In the search process, we consider a total of $6$ cells with the initial $16$ channels  in the network, where the reduction cell are inserted in the second and the fourth layers, and the others are normal cells. There are $M=4$ intermediate nodes in each cell. We set $T = 3$ and the initial number of operations $K$ is set to $8$, so the final number of search epochs is $(8+7+6+5+4+3+2)*3=105$. $\beta_P$ is set as $2$, and the batch size is set to $512$. We use SGD with momentum to optimize the network weights, with an initial learning rate of $0.025$ (annealed down to zero following a cosine schedule), a momentum of 0.9, and a weight decay of $5 \times 10^{-4}$. When we search for the architecture directly on ImageNet, we use the same parameters for searching with CIFAR-10 except that the initial learning rate is set to $0.05$ and $\beta_P$ is set to $0.33$. Due to the efficient guidance of CP model, we only use $50$\% of the training set with CIFAR-10 and ImageNet for architecture search \zhuo{and $5$\% of the training set for evaluation,} leading to a faster search.
	
	After search, in the architecture evaluation step, our experimental settings are similar to \cite{liu2018darts,Zoph2018CVPR,pham2018efficient}. A larger network of $10$ cells ($8$ normal cells and $2$ reduction cells) is trained on CIFAR-10 for $600$ epochs with a batch size of $96$ and an additional regularization cutout \cite{devries2017improved}. The initial number of channels is set as $56$, $72$, $112$ for different model sizes. We use the SGD optimizer with an initial learning rate of $0.025$ (annealed down to zero following a cosine schedule without restart), a momentum of $0.9$, a weight decay of $3 \times 10^{-4}$ and a gradient clipping at $5$. When stacking the cells to evaluate on ImageNet, the evaluation stage follows that of DARTS \cite{liu2018darts}, which starts with three convolutional layers with a stride of  $2$ to reduce the input image resolution from $224 \times 224$ to $28 \times 28$. $10$ cells ($8$ normal cells and $2$ reduction cells) are stacked after these three layers, with the initial channel number being $102$. The network is trained from scratch for $250$ epochs using a batch size of $256$. We use the SGD optimizer with a momentum of $0.9$, an initial learning rate of $0.05$ (decayed down to zero following a cosine schedule), and a weight decay of $3 \times 10^{-5}$. Additional enhancements are adopted including label smoothing and an auxiliary loss tower during training. All the experiments and models are implemented in PyTorch \cite{paszke2017automatic}.
	
	\begin{table*}
		\centering
		\footnotesize
		\begin{tabular}{lccccccc}
			\toprule
			\multirow{2}{*}{\textbf{Architecture}} & \multicolumn{2}{c}{\textbf{Accuracy (\%)}} & \textbf{Params}& \textbf{W}&\textbf{A} & \textbf{Search Cost} & \textbf{Search} \\ \cline{2-3}
			& \textbf{Top1} & \textbf{Top5} & \textbf{(M)} &&& \textbf{(GPU days)} & \textbf{Method} \\ 
			\hline
			ResNet-18 \cite{gu2018projection} & 69.3 & 89.2 & 11.17 &32&32 & - & Manual \\
			PNAS \cite{liu2018progressive} & 74.2 & 91.9 & 5.1 &32&32 & 225 & SMBO \\
			DARTS \cite{liu2018darts} & 73.1 & 91.0 & 4.9 &32&32 & 4 & Gradient-based \\
			PC-DARTS \cite{xu2019pcdarts} & 75.8 & 92.7 & 5.3 &32&32 & 3.8 & Gradient-based \\
			
			\hline
			ResNet-18 (PCNN) \cite{gu2018projection} & 63.5 & 85.1 & 11.17 &1&32 & - & Manual \\
			BNAS (PCNN) \cite{chen2019BNAS} & 71.3 & 90.3 & 6.2 &1&32 & 2.6 & Performance-based \\					
			\hline
			ResNet-18 (Bi-real Net) \cite{liu2018bi} & 56.4 & 79.3 & 11.17 &1&1 & - & Manual \\
			
			ResNet-18 (PCNN) \cite{gu2018projection} & 57.5 & 80.0 & 11.17 &1&1 & - & Manual \\
			ResNet-18 (BONN) \cite{zhao2019bonn} & 59.3 & 81.5 & 11.17 &1&1 & - & Manual \\		
			\textbf{CP-NAS*}  & \textbf{64.3} & 85.6 & 12.5 &1&1 & 2.8 & Child-Parent model \\								\textbf{CP-NAS**}  & \textbf{66.5} & 86.8 & 12.5 &1&1 & 2.8 & Child-Parent model \\						
			\bottomrule
			\bottomrule[1pt]
		\end{tabular}
		\caption{Test accuracy on ImageNet. * represents that we do not binarize the activations of the first convolutional layer in depth-wise separable convolution. ** represents that we do not binarize the activations of preprocessing operations for $2$ input nodes either.}
		\label{tab:imagenet}
	\end{table*}
	
	\subsection{Results on CIFAR-10}
	We first evaluate our CP-NAS on CIFAR-10 and compare results with both manually designed networks~\cite{zagoruyko2016wide} and networks searched by NAS~\cite{liu2018darts,xu2019pcdarts} at different levels of binarization.
	
	\begin{figure}[bp!]
		\centering
		\subfigure[The normal cell]{
			\label{Train Error}
			\includegraphics[scale=0.33]{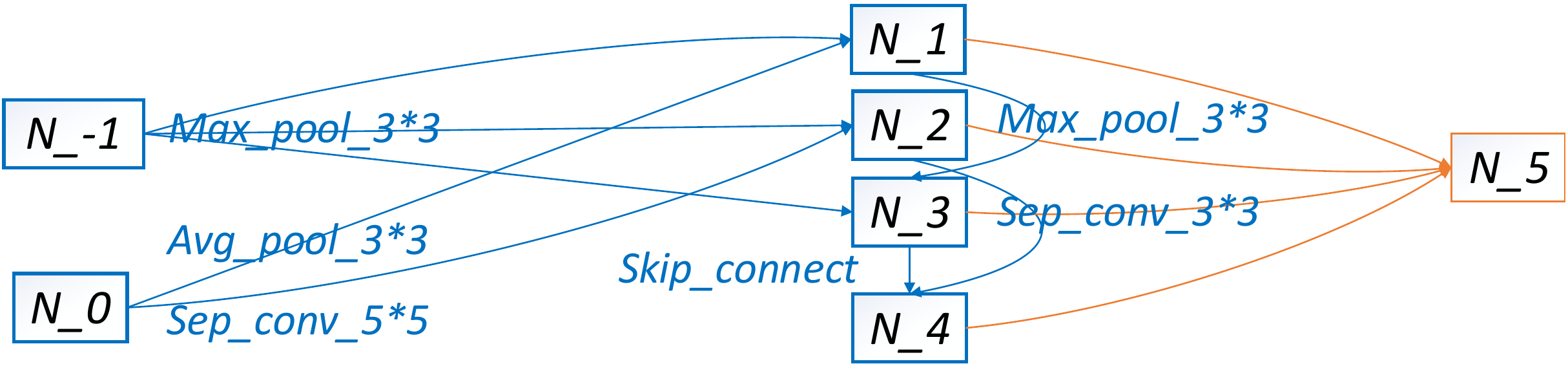} }
		\subfigure[The reduction cell]{
			\label{Test Error}
			\includegraphics[scale=0.33]{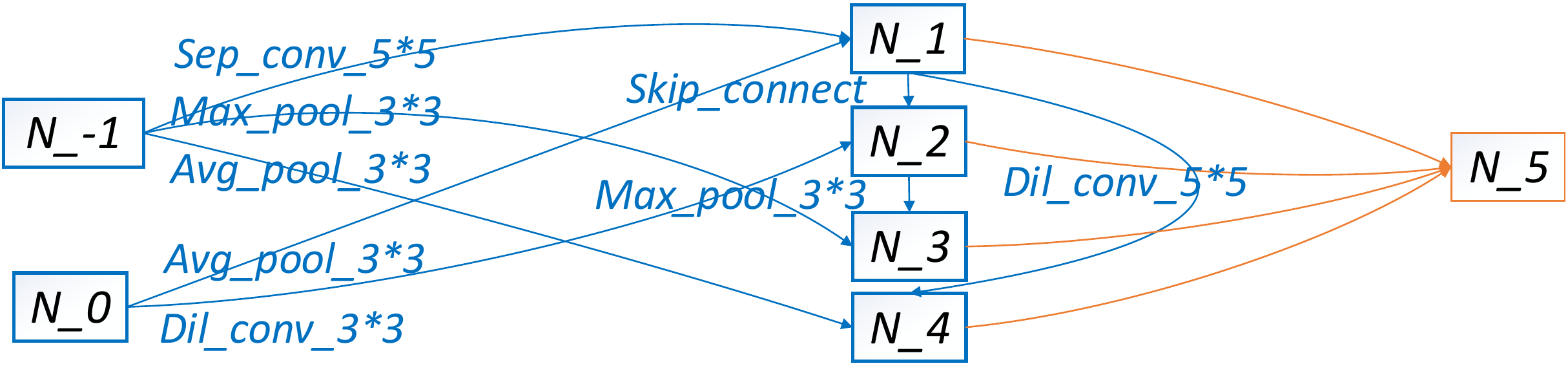} }
		\caption{The normal cell (a) and the reduction cell (b) searched for CIFAR-10.}
		\label{fig:normal_reduction}
	\end{figure}		
	
	The results for different architectures on CIFAR-10 are summarized in Tab.~\ref{tab:cifar}. 
	We search for three binarized networks with different model sizes which binarize both weight and activation. Note that for the model size, in addition to the number of parameters, we should also consider the number of bits of each parameter. The binarized networks only need $1$ bit to save and compute the weight parameter or the activation parameter, while the full-precision networks need $32$. \zhuo{More details about the efficiency are discussed in section 3.4.}
	
	Compared with manually designed networks, e.g., WRN-22(BONN) \cite{zhao2019bonn}, our CP-NAS achieved comparable or smaller test errors { ($6.5$\% vs. $8.07$\%)} and more compressed models  { ($2.9$M vs. $4.33$M)}. Compared with full-precision networks obtained by other NAS methods, our CP-NAS achieved comparable test errors and significantly more compressed models, with similar or less search time. Compared with BNAS, our CP-NAS binarized the activation parameters and achieved comparable test errors with only slightly longer search time. \zhuonew{We further implement BNAS$^\dagger$ for 1-bit CNNs by setting $\beta_P=0$ in CP-NAS, which means that we only use the performance measure for the operation selection. The result shows that CP-NAS achieve better performance than BNAS$^\dagger$ with lower test error { ($5.72$\% vs. $8.29$\%)} and similar model size. CP-NAS outperforms BNAS in both network efficiency and 1-bit CNNs performance.} More detailed comparison with BNAS is presented in section 3.4.
	
	In terms of search efficiency, compared with the previous work PC-DARTS \cite{xu2019pcdarts}, our CP-NAS is $30\%$ faster (tested on our platform - 6 NVIDIA TITAN V GPUs). We attribute our superior results to the proposed scheme of search space reduction. As shown in Fig.~\ref{fig:normal_reduction}, the architectures of CP-NAS prefer smaller receptive fields. Our CP-NAS also results in more pooling operations, which can increase the nonlinear representation ability of BNNs.

	\subsection{Results on ImageNet}
	To further evaluate the performance of our CP-NAS, we compare our method with the state-of-the-art image classification methods on the ImageNet. All the searched networks are obtained directly by CP-NAS on ImageNet by stacking the cells. Due to the first convolutional layer in a depth-wise separable convolution \zhuo{with fewer parameters,} we do not binarize the activations of the first layer for ImageNet. Tab.~\ref{tab:imagenet} shows the test accuracy on ImageNet. We observe that CP-NAS outperforms manually designed binarized networks { ($64.3$\% vs. $59.5$\%)} with a similar number of parameters { ($12.5$M vs. $11.17$M)}. 
	Note that compared to the human-designed full-precision networks, our CP-NAS achieved comparable performance but with higher compression. 
	Furthermore, to obtain a better performance, we do not binarize the activations of the preprocessing operations for the two input nodes, and achieve an accuracy of $66.5$\%, which is much closer to the full-precision hand-crafted model, e.g., $69.3$\% for ResNet-18. 

	\begin{figure}[bp!]
		\centering
		\includegraphics[scale=.33]{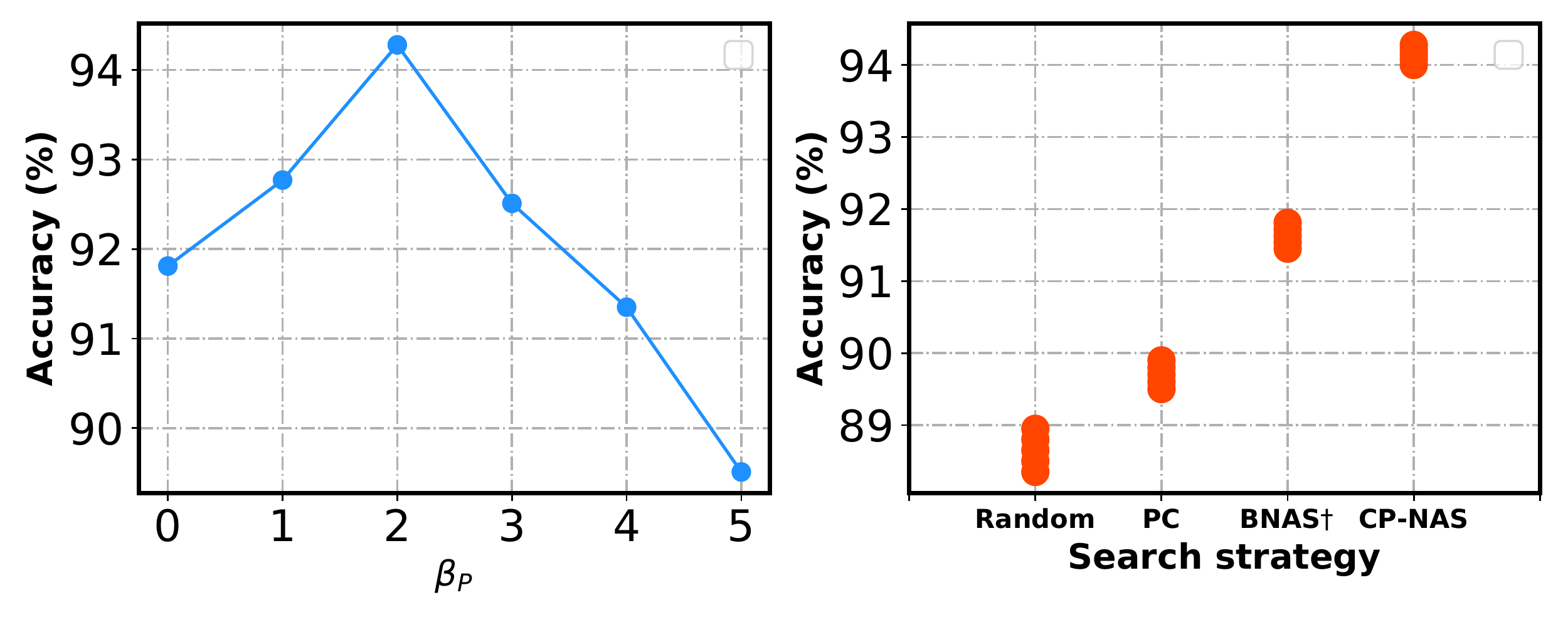}
		\caption{The result (left) for different $\beta_P$ on CIFAR-10. The 1-bit CNNs result (right) for different search strategys on CIFAR-10, including Random (Random selection), PC (PC-DARTs), BNAS$^\dagger$, CP-NAS.}
		\label{fig:search}
	\end{figure}

	\subsection{Ablation Study}
	We test different $\beta_P$ for our method on the  CIFAR-10 dataset, as shown in Fig.~\ref{fig:search}. We can see that when  $\beta_P$ increases, the accuracy increases at the beginning, but decreases when $\beta_P \ge 2$. It validates that the performance loss between the Child and Parent models is a significant measure for 1-bit CNNs search. When $\beta_P$ becomes larger, CP-NAS tends to select the architecture with fewer convolutional operations, but  a large imbalance between two elements in our CP model will cause a performance drop.   
	
	We also compare the architectures obtained by CP-NAS, Random (Random selection), PC (PC-DARTs) and BNAS$^\dagger$ as shown in Fig.~\ref{fig:search}. Unlike the case of the full-precision model, Random and PC-DARTs lack the necessary guidance, which have a poor performance for binarized architecture search. Both BNAS$^\dagger$ and CP-NAS have the evaluation indicator for the operation selection. Differently, our CP-NAS also considers an additional performance loss, which can outperform the other three strategies.
	
	\begin{table}[tbp!]
		\centering
		\scriptsize
		\begin{tabular}{lcccc}
			\toprule[1pt]
			{\textbf{Architecture}} & \textbf{Memory usage} & \textbf{Memory} & \textbf{FLOPs}&\textbf{Speedup$\uparrow$} \\
			& \textbf{(Mbits)$\downarrow$} & \textbf{saving$\uparrow$} & \textbf{(M)$\downarrow$} &  \\
			\hline
			
			WRN-22  & 138.27 & 1$\times$ & 647.70 & 1$\times$  \\			
			\hline						
			BNAS (PCNN, larger) & $\approx$5.12 & $\approx$27$\times$  & $\geq$300 & $\leq$2.16$\times$ \\ 
			\hline
			WRN-22 (BONN) & 5.71 & 24.19$\times$ &17.03&28.03$\times$   \\				
			\textbf{CP-NAS} (Small) & 3.32 & 41.56$\times$ & 12.89 & 50.24$\times$ \\
			\textbf{CP-NAS} (Medium) & 4.85 & 27.93$\times$ & 18.30 & 35.39$\times$ \\
			\textbf{CP-NAS} (large) & 11.51 & 12.01$\times$ & 38.67 & 16.75$\times$ \\ 		
			\bottomrule[1pt]
		\end{tabular}
		\caption{\zhuonew{Comparison of memory saving and speedup of BNAS (PCNN, larger), WRN-22 (BONN) and  CP-NASs (Small, Medium, Large) on CIFAR-10 with respect to WRN-22. $\uparrow$ represents that the larger is better, vice versa for $\downarrow$.}}
		\label{tab:memory_speed}
	\end{table}
	
	\zhuo{\textbf{Efficiency.}  The 1-bit CNNs \zhuonew{are extremely} efficient for resource-limited devices, showing up to 58$\times$ speedup and 32$\times$ memory saving \zhuonew{than the full-precision models} \cite{rastegari2016xnor}. As shown in Tab.~\ref{tab:memory_speed}, our CP-NASs (Small, Medium, Large) for CIFAR-10 achieve comparable performance as the full-precision hand-crafted WRN-22 model, with $41.56\times$, $27.93\times$, $12.01\times$ memory saving and $50.24\times$, $35.39\times$, $16.75\times$ speedup in terms of FLOPs. \zhuonew{As a result, our CP-NAS models bring significant benefits for resource-contrained edge computing appliations.}}
	
	\section{Conclusion}
	In this paper, we  calculate 1-bit CNNs  based on the proposed Child-Parent  model under the full-precision network supervision.  We build a bridge between 1-bit CNNs and NAS using our proposed CP model, leading to the CP-NAS method. 
	With our proposed CP-NAS, we are able to solve the  neural architecture search and the binarized optimization in the same framework. 
	Experiments on CIFAR-10 and ImageNet datasets demonstrate that our method achieves better performance than other state-of-the-art methods  with a  more compressed model and less search time. The future work will focus on more applications, such as object detection \zhuonew{and tracking}.
	
	\section*{Acknowledgements}
	Baochang Zhang is also with Shenzhen Academy of 
	Aeros-pace Technology, Shenzhen, China, 
	and he is the corresponding author.  
	He is in part Supported by National Natural Science Foundation of China under Grant 61672079, Shenzhen Science and Technology Program (No.KQTD20-16112515134654).

	\bibliographystyle{named}
	\bibliography{ijcai20}
	
\end{document}